\documentclass{article}

\usepackage{PRIMEarxiv}

\usepackage[utf8]{inputenc} % allow utf-8 input
\usepackage[T1]{fontenc}    % use 8-bit T1 fonts
\usepackage{hyperref}       % hyperlinks
\usepackage{url}            % simple URL typesetting
\usepackage{booktabs}       % professional-quality tables
\usepackage{amsfonts}       % blackboard math symbols
\usepackage{nicefrac}       % compact symbols for 1/2, etc.
\usepackage{microtype}      % microtypography
\usepackage{lipsum}
\usepackage{fancyhdr}       % header
\usepackage{graphicx}       % graphics
\graphicspath{{media/}}     % organize your images and other figures under media/ folder

\usepackage{amsmath} % assumes amsmath package installed
\usepackage{multirow} % 
\usepackage{booktabs}
\usepackage{graphicx}
\usepackage{xcolor}
\usepackage{caption}
\usepackage{makecell}
\usepackage{comment}
\usepackage{hyperref} 
\urldef{\codeurl}\url{https://github.com/GuangxunZhu/VehCondPose3D}

\newcommand{\GX}[1]{}%\textcolor{blue}{[GX: #1]}}

\newcommand{\hi}[1]{\textbf{#1}}

\title{Modeling 3D Pedestrian-Vehicle Interactions for Vehicle-Conditioned Pose Forecasting
}

\author{Guangxun Zhu\thanks{School of Computing Science, University of Glasgow, Glasgow, United Kingdom} \hspace{0.5cm}Xuan Liu\footnotemark[1] \hspace{0.5cm}Nicolas Pugeault\footnotemark[1] \hspace{0.5cm}Chongfeng Wei\thanks{James Watt School of Engineering, University of Glasgow, Glasgow, United Kingdom} \hspace{0.5cm}Edmond S. L. Ho\footnotemark[1] \thanks{Corresponding author {\tt\small Shu-Lim.Ho@glasgow.ac.uk}}% <-this % stops a space
}

\begin{document}
\maketitle

\begin{abstract}
Accurately predicting pedestrian motion is crucial for safe and reliable autonomous driving in complex urban environments. In this work, we present a 3D vehicle-conditioned pedestrian pose forecasting framework that explicitly incorporates surrounding vehicle information. To support this, we enhance the Waymo-3DSkelMo dataset with aligned 3D vehicle bounding boxes, enabling realistic modeling of multi-agent pedestrian–vehicle interactions. We introduce a sampling scheme to categorize scenes by pedestrian and vehicle count, facilitating training across varying interaction complexities. Our proposed network adapts the TBIFormer architecture with a dedicated vehicle encoder and pedestrian–vehicle interaction cross-attention module to fuse pedestrian and vehicle features, allowing predictions to be conditioned on both historical pedestrian motion and surrounding vehicles. Extensive experiments demonstrate substantial improvements in forecasting accuracy and validate different approaches for modeling pedestrian–vehicle interactions, highlighting the importance of vehicle-aware 3D pose prediction for autonomous driving. Code is available at: \codeurl
\end{abstract}

% keywords can be removed

\section{Introduction}
%~\todo{only 22 refs, looks like I need to add a little more. Questions:Is the term vehicle-conditioned appropriate?} \eh{Yes, we need to discuss those - is there any 3D skeletal motion + vehicle dataset in the literature?}
% A comprehensive understanding of the interactions between intelligent agents (e.g., pedestrians and vehicles) and their surrounding environment is fundamental for enabling autonomous vehicles to attain robust perception and safe planning in complex urban environments. Recent studies have demonstrated that incorporating interaction-awareness into perception systems yields substantial advancements in tasks such as pedestrian trajectory prediction~\cite{SocialCircle} and 3D pose forecasting~\cite{peng2023trajectory,xiao2024multi}.

% \todo{1. confirm 1-2 or 1-3 scenes and modify the expression of relevant parts;2. Evaluation results and relevant expression;3. Ablation study results and expression;4. Qualitative comparison}
Autonomous driving has attracted significant attention for its potential to revolutionize transportation, but it still faces numerous challenges in complex urban environments~\cite{wang2024driving}. Among these, pedestrian prediction is particularly critical, as pedestrians are common, vulnerable, and highly dynamic road users, whose flexible and sometimes unpredictable behaviors pose significant challenges for perception systems~\cite{Ham_2023_CVPR}. Accurately understanding these behaviors is therefore essential for safe navigation and planning.

% A central challenge in interaction modeling lies in accurately capturing the fine-grained dynamics of pedestrian behavior. As critical participants in road environments, pedestrians require particular consideration in the design of autonomous driving systems. Their behavior is characterized by active collision avoidance with other road users (e.g., pedestrians, vehicles, cyclists) while simultaneously being constrained by roads, traffic regulations, and surrounding infrastructure. Inadequate or overly coarse modeling of these dynamics can result in erroneous predictions of pedestrian trajectories, potentially leading autonomous vehicles to adopt inappropriate responses to the situation.
% \todo{I will add more related works...}

A main challenge in pedestrian prediction lies in modeling their interactions with other agents. Pedestrian behavior is influenced by other road users (e.g., pedestrians, vehicles, cyclists)~\cite{Crosato:review}, and insufficient or overly coarse modeling of these interactions can result in inaccurate trajectory predictions, potentially leading autonomous vehicles to adopt inappropriate responses. 
Most prior studies~\cite{SocialCircle,chen2025socialmoif,yao2024trajclip, Crosato:ICHMS2021, Crosato:TIV2023, PedFormer,azarmi2024pip} model pedestrian interactions using 2D representations such as locations and 2D poses. For handling data captured from the bird's-eye view,  SocialCircle~\cite{SocialCircle} introduces an angle-based representation to capture relative spatial layouts for interaction modeling, and SocialMOIF~\cite{chen2025socialmoif} leverages multi-order intention fusion to enhance trajectory prediction. TrajCLIP~\cite{yao2024trajclip} proposed to improve the continuity in the predicted pedestrian trajectories using contrastive learning to improve the consistency between the feature spaces of the historical and future trajectories. Social Value Orientation (SVO), which is a concept in Psychology for a person's preferences for allocating resources between themselves and others, has been incorporated into Reinforcement Learning (RL) frameworks~\cite{Crosato:ICHMS2021,Crosato:TIV2023} for modeling different behaviours between pedestrians and drivers. While these methods explicitly consider social interactions, they remain limited to 2D numerical coordinates, which restricts their ability to capture fine-grained 3D motion dynamics.

\begin{figure} [!htb]
    \centering
    \includegraphics[width=\columnwidth]{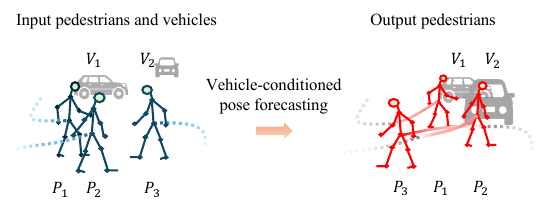}
    \caption{Illustration of our Vehicle-conditioned pedestrian pose forecasting. Pedestrian predictions are not only based on their historical motion but are also influenced by surrounding vehicle information.}
    \label{fig:pose forecasting}
\end{figure}

%\eh{I try to add some work related to ego-centric view here, maybe we can talk about bird-eye view ped-veh interaction work in the previous paragraph?} 
Another stream of existing research in pedestrian-vehicle interactions focused on analysing ego-centric view data available in public datasets such as TITAN~\cite{malla2020titan}, JAAD~\cite{rasouli2017ICCVW,Rasouli2017IV} and PIE~\cite{Rasouli2019PIE}. A recent work, namely PedVLM~\cite{Munir:VLM:2025}, predicts the intention (i.e. crossing or not crossing) of the pedestrian by feeding the RGB image and the corresponding optical flow computed from the ego-centric view alongside the textual description of the scene to a large visual-language
models (VLM). PedFormer~\cite{PedFormer} fuse multimodal features, including 2D coordinates of pedestrians from the ego-centric view, ego-vehicle motion and semantic segmentation of the scene for pedestrian behavior prediction. Further to multimodal sensor fusion, PIP-Net~\cite{azarmi2024pip} takes advantage of fusing features extracted from multi-camera views (left,
front, and right) captured using on-board cameras provided in the new Urban-PIP dataset to better model the contextual information when modeling pedestrian-vehicle interactions. 

Although some approaches may leverage a combination of multiple modalities, such as images, vehicle ego-motion, 2D pedestrian poses, or vehicle 2D bounding boxes, they are limited in capturing fine-grained pedestrian behaviors and motion dynamics, which may result in inaccurate predictions and unsafe planning decisions for autonomous vehicles. More recently, several works have explored the use of 3D human poses to better capture motion dynamics~\cite{Distillation, saadatnejad2023social}. Compared to 2D data, 3D poses provide richer spatial structures and depth information, thereby providing a stronger basis for fine-grained interaction modeling. Nevertheless, these approaches focus only on pedestrian–pedestrian interactions and do not consider the influence of multiple types of agents in autonomous driving scenarios, primarily due to the scarcity of 3D pose datasets captured in real driving environments. For example, many works rely on MuPoTS-3D~\cite{MuPoTs-3D}, which provides 3D poses estimated only from RGB images and is collected in non-driving contexts, limiting realism. Wang et al.~\cite{Wang:NeurIPS2021} attempted to synthesize three-person interactions by combining single- and two-person motion sequences from the high-quality CMU-Mocap dataset~\cite{mocapdata}, recorded using an optical motion capture system. However, this produces a synthetic dataset with limited diversity and lacks realistic multi-agent interactions in traffic scenarios. The most recent dataset, JRDB-GlobMultiPose (JRDB-GMP)~\cite{Jeong:CVPR2024}, is collected in real-world environments using a moving robot platform, yet the 3D poses are obtained purely via RGB-based pose estimation,\GX{which affetcs the motion quality due to depth ambiguity and reduced robustness under occlusion.}  Moreover, these datasets lack aligned 3D data from real autonomous driving scenarios, making it difficult for models trained on them to generalize to realistic interactions in such environments.

% To address the challenges of data scarcity and improve data quality, the Waymo-3DSkelMo~\cite{Waymo-3DSkelMo} is constructed from the Waymo Open Dataset Perception Benchmark (hereafter referred to as Waymo)~\cite{waymo_open_dataset}. Specifically, 3D human shape (estimated using LiDAR-HMR~\cite{fan2023lidar} with SMPL~\cite{loper2023smpl}) and motion~\cite{NeMF} priors were used to reconstruct high-quality and natural 3D skeletal motion from the raw LiDRA range images in the Waymo dataset. While the Waymo-3DSkelMo significantly boosted the number of 3D skeletal poses from just under 9,976 in Waymo to over 2.4 million, only 3D skeletal motion data is available and benchmarked which limits the usage of the dataset in autonomous driving applications. \todo{Add more related work on pedestrian-vehicle and the datasets}     

To address this challenge, Crosato et al.~\cite{Crosato:VR} proposed a VR platform for capturing interactions between a pedestrian and a vehicle controlled by human subjects in 3D in a virtual environment. However, the data was not captured in real-world and only contains limited scenarios with 1 pedestrian and 1 vehicle. More recently, Waymo-3DSkelMo~\cite{Waymo-3DSkelMo} was constructed from the Waymo Open Dataset Perception Benchmark (hereafter referred to as Waymo)~\cite{waymo_open_dataset}, a large-scale autonomous driving dataset, to provide 3D pedestrian skeletal motion data. Specifically, 3D human shapes (estimated using LiDAR-HMR~\cite{fan2023lidar} with SMPL~\cite{loper2023smpl}) and motion priors (Neural Motion Fields (NeMF)~\cite{NeMF}) were used to reconstruct high-quality and natural 3D skeletal motion from the raw LiDAR range images in Waymo. Waymo-3DSkelMo significantly increases the number of 3D skeletal poses from approximately 10k in Waymo to over 2.4 million and is aligned with other data modalities, such as 3D vehicle bounding boxes and LiDAR point clouds, making it possible to model full-scene interactions. However, only 3D skeletal motion is benchmarked in Waymo-3DSkelMo, which limits its usage for broader autonomous driving applications.

In this paper, we extend the Waymo-3DSkelMo dataset by incorporating the 3D bounding box information of each vehicle from the Waymo dataset\GX{, since vehicles are the most common and safety-critical interacting agents that strongly influence pedestrians}. Through aligning the vehicle and skeletal motion in a common 3D space, 3D data representing the interactions between multiple pedestrians and vehicles can be obtained. We further propose a sampling scheme to divide the scenes into different categories to facilitate the training of interaction modeling networks with different complexity levels (i.e. different numbers of vehicles and pedestrians). Finally, we propose a 3D Vehicle-conditioned pedestrian pose forecasting network by incorporating the vehicle information and adapting the TBIFormer~\cite{peng2023trajectory} architecture.
%Finally, we propose a 3D pedestrian pose forecasting network by incorporating vehicle information and adapting the TBIFormer~\cite{peng2023trajectory} architecture. 
In this network, pedestrian predictions not only rely on their historical motion but are also conditioned on surrounding vehicle information, as illustrated in Fig.~\ref{fig:pose forecasting}. Extensive experimental results are obtained to demonstrate the benefits of including the vehicle information in the 3D pose forecasting of the pedestrians.

Our contributions can be summarized as follows:

% \begin{itemize}
%   \item ... enhanced the Waymo-3DSkelMo dataset 
%   \item ... new 3D pedestrian pose forecasting network with vehicle information
%   \item ... benchmarking results
%   \item ... open source code and data
% \end{itemize}

\begin{itemize}
  \item We enhance the Waymo-3DSkelMo dataset by incorporating 3D vehicle bounding boxes and introducing a scene-level sampling scheme that categorizes interactions based on the number of pedestrians and vehicles, enabling more realistic and structured modeling of pedestrian–vehicle interactions.
  \item We propose a new 3D pedestrian pose forecasting network that incorporates vehicle information, allowing pedestrian predictions to be conditioned on both historical pedestrian motion and surrounding vehicles.
  \item We provide extensive benchmarking results, demonstrating the benefits of incorporating vehicle information for accurate 3D pose forecasting, and validating different approaches to modeling interactions between pedestrians and vehicles. \textit{The enhanced dataset, experimental protocol (e.g. data split) and code will be available.}
  % \item ... open source code and data
\end{itemize}

\section{Dataset} \label{sec:dataset}
In this section, we first introduce the enhancement of the Waymo-3DSkelMo~\cite{Waymo-3DSkelMo} dataset by incorporating vehicle information in Section~\ref{sec:veh_info}. %Next, ....%we present our proposed network for modelling the vehicle and pedestrian information in 3D (Section ??).
Next, we present the scene segmentation strategy in Section~\ref{sec:scene_seg}, which is designed to sample diverse data for the experiments.

\subsection{Incorporating Vehicle Information} \label{sec:veh_info}

We employ Waymo-3DSkelMo~\cite{Waymo-3DSkelMo} and Waymo Open Dataset (Perception)~\cite{waymo_open_dataset} to train and validate pedestrian–vehicle interaction models. In particular, Waymo-3DSkelMo contains 2,438,145 3D skeletal poses of the pedestrian reconstructed from the raw LiDAR range images provided by the Waymo dataset. The naturalness of the reconstructed pedestrian motions was further enhanced using NeMF~\cite{NeMF} as the human motion prior. Waymo-3DSkelMo provides 3D keypoints of all pedestrians in real-world scenes, which can be spatially and temporally aligned with the vehicle 3D bounding boxes in Waymo in this %, making them highly compatible for our 
study. %In total, 
Waymo-3DSkelMo contains \textbf{837 scenes} with an overall duration of %\textbf{14,419 seconds}, 
\textbf{4 hours}, covering a diverse set of urban scenarios and capturing a large number of pedestrian–vehicle interactions.

\subsection{Scene Segmentation} \label{sec:scene_seg}
%However, as 
Because different scenes contain varying numbers of agents with diverse spatial and temporal distributions, %distributions and temporal overlaps, 
it is difficult to utilize all scenes as a single training dataset due to the large difference in spatial density. %the data directly. 
As a result, further segmentation of the scenes %data filtering 
is necessary to provide more meaningful training data for downstream tasks. %accommodate the model input. 

% We segmented the scenes based on the proximity between pedestrians and vehicles nearby. Specifically, \todo{explain in detail how to do the KDtress...} In order to segment the scenes As illustrated in Fig.~\ref{fig:dataset_filtering}, we first apply a KDTree~\cite{bentley1975multidimensional} nearest-neighbor to extract candidate scenes containing fixed numbers of pedestrians %(e.g., one-person and two-person scenes) 
% from all available data. 

\begin{figure}[htb]
  \centering
  \includegraphics[width=\columnwidth]{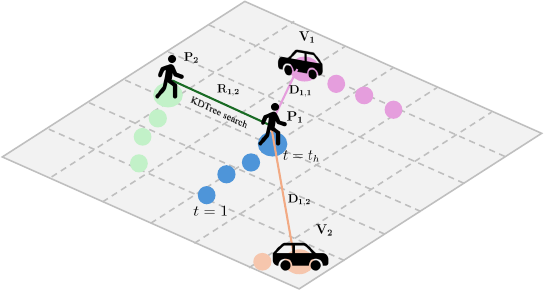}
  \caption{Illustration of pedestrians and vehicles in a scene with their trajectories and inter-agent distances.}
  \label{fig:dataset_filtering}
\end{figure}

In particular, we segmented the scenes based on the distance between pedestrians and vehicles nearby. Specifically, as illustrated in Fig.~\ref{fig:dataset_filtering}, we first apply a KDTree~\cite{bentley1975multidimensional}, a data structure that enables efficient nearest-neighbor searches in multi-dimensional space, to quickly find pedestrian groups that are closest to each other based on the average root joint positions over a temporal window. In this work, we focus on the extracted scenes with the number of pedestrians between 1 and 3 since the number of scenes decreases sharply above this range as indicated in the statistics presented in Table~\ref{tab:num_samples}. %\todo{to confirm if we focus on 1-3} 
A sliding window is then employed to select temporally overlapping segments. %\eh{Before this, we need to justify why we pick 1- and 2- person scenes. We can show a table of statistics on the number of scenes with different numbers of pedestrians like those you showed me before?} 
% For each segment, we check whether the inter-pedestrian distance $R$ at any timestamp falls below a threshold of 18 m, which follows the maximum distance used in the TBIFormer~\cite{peng2023trajectory} training dataset CMU-Mocap (UMPM)~\cite{mocapdata, UMPM}.
For each segment, we compute the maximum pairwise distance between pedestrians' root positions at each timestamp and then take the minimum over all timestamps, denoted as $R$:% (see Eq.~\ref{eq:min_max_distance}). 
\begin{equation}
R = \min_{t=1,\dots,T} \Big( \max_{i,j \in \{1,\dots,N_p\}, i\neq j} \|\mathbf{r}_i(t) - \mathbf{r}_j(t)\|_2 \Big)
\label{eq:min_max_distance}
\end{equation}
Here, $\mathbf{r}_i(t)$ represents the 3D root position of the $i$-th pedestrian at time $t$, $N_p$ is the number of pedestrians in the segment, and $T$ is the total number of frames in a time window. We retain the segment if $R$ is below a threshold of 18~m which is the maximum distance used in the TBIFormer~\cite{peng2023trajectory} training dataset CMU-Mocap (UMPM)~\cite{mocapdata, UMPM}.

\begin{table}
\caption{Number of segmented scenes (training + validation) with different numbers of pedestrians and vehicles within 0–15~m in the dataset.}
\label{tab:num_samples}
\centering
\begin{tabular}{ c c c c c }
\hline
\multirow{2}{*}{\thead{\# of\\Veh.}} & \multicolumn{4}{c}{Number of Pedestrian(s)}\\ 
\cmidrule(lr){2-5}
 & 1 & 2 & 3 & 4\\
\hline
% just training
% 0 & 17762 & 8236 & 5160 & 3503\\ 
% 1 & 15446 & 7534 & 4975 & 3541\\  
% 2 & 14336 & 7321 & 5077 & 3810\\
% 3 & 12982 & 6815 & 4903 & 3764\\
% 4 & 10683 & 6192 & 4351 & 3393\\

% all
0 & 17762 + 8236 & 8236 + 2232 & 5160 + 1333 & 3503 + 875\\ 
1 & 15446 + 7534 & 7534 + 2005 & 4975 + 1355 & 3541 + 977\\  
2 & 14336 + 7321 & 7321 + 1782 & 5077 + 1233 & 3810 + 883\\
3 & 12982 + 6815 & 6815 + 1733 & 4903 + 1245 & 3764 + 902\\
4 & 10683 + 6192 & 6192 + 1272 & 4351 + 944 & 3393 + 738\\

\hline
\end{tabular}
\end{table}

\begin{figure}[htb]
  \centering
  \includegraphics[width=\columnwidth]{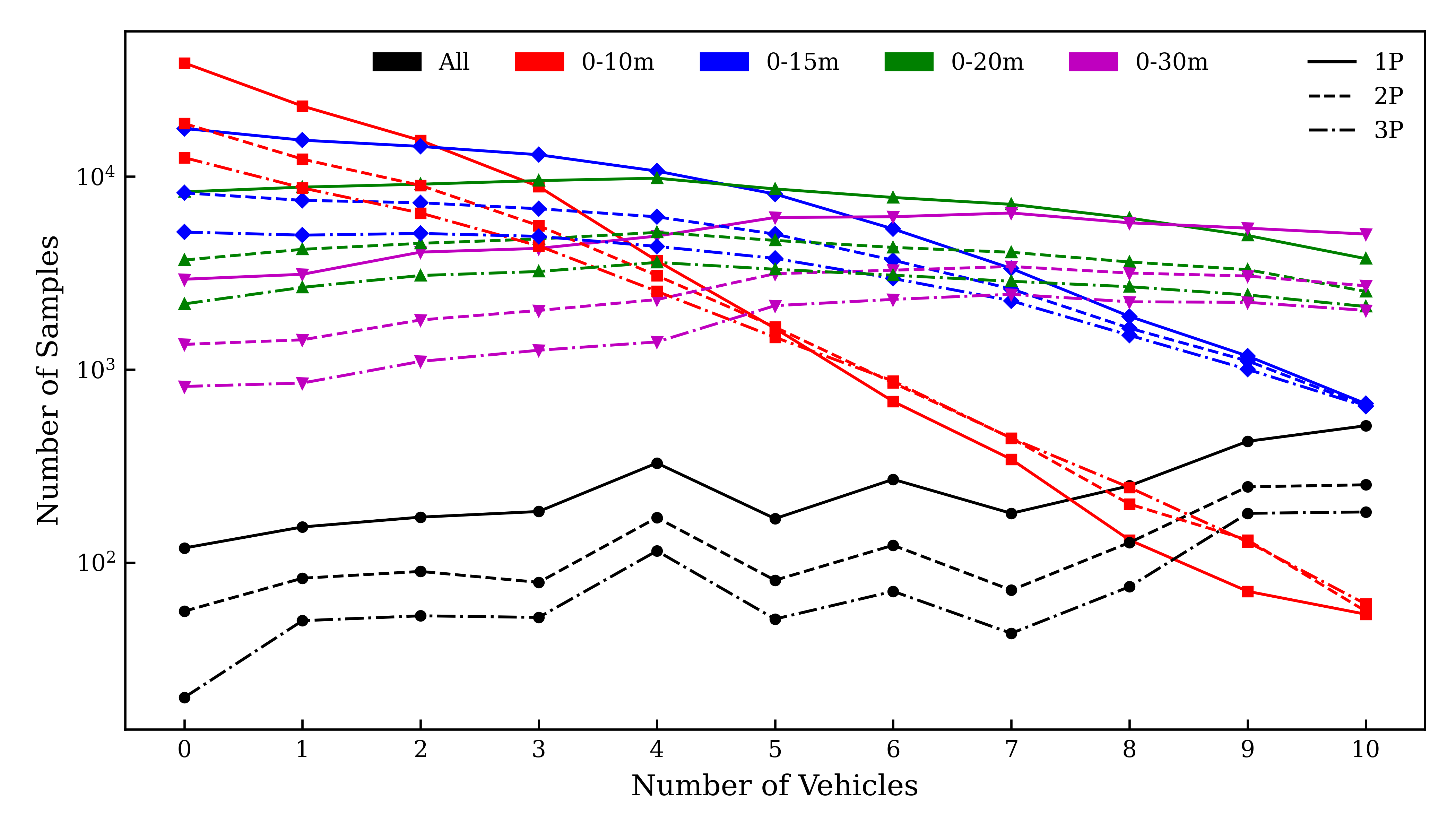}
  \caption{Number of pedestrian–vehicle interaction training samples under varying numbers of surrounding vehicles and different distance thresholds for scenarios ranging from one to three pedestrians.}
  \label{fig:dataset_statistics}
\end{figure}

Since the number of surrounding vehicles varies across pedestrian samples, this inconsistency introduces challenges for training and validation. To address this, %we set a pedestrian–vehicle distance threshold and select all vehicles within this range, where the distance is measured as the average distance between each vehicle and at least one pedestrian over the overlapping frames. 
%\GX{
we select the set of vehicles $V_{\text{selected}}$ whose average distance to the nearest pedestrian over the frames in the time window $T$ is below a predefined threshold $th$:
\[
V_{\text{selected}} = \left\{ i \in V \;\middle|\; \frac{1}{|T|} \sum_{t \in T} \min_{j \in P} \| r_j(t) - v_i(t) \|_2 \le th \right\}
\]
where $r_j(t)$ and $v_i(t)$ denote the positions of pedestrian $j$ and vehicle $i$ at frame $t$, respectively.

As shown in Fig.~\ref{fig:dataset_statistics}, the number of training samples associated with different vehicle counts varies under different distance thresholds. %In the one-person case, we obtain 115,278 samples in total (92,921 for training), while the two-person case yields 64,614 samples (52,229 for training) 
%\todo{add 3-ped statistics}.\GX{I think we can remove these statistics, because these are obtained only by filtering the distance between pedestrians, not including vehicles. The numbers considering vehicles of different ranges are shown in the figure~\ref{fig:dataset_statistics}. I am afraid that this may cause misunderstanding. The data we use are shown in the table~\ref{tab:num_samples}, which we can refer to later in the paragraph.} %To balance data diversity and sample availability, we retain samples within a distance range of 0–15 m, which cover the majority of samples and yield a relatively balanced distribution.
From Fig.~\ref{fig:dataset_statistics}, we can see that if the distance threshold is too small, the number of available samples decreases sharply as the number of vehicles increases. Conversely, if the threshold is too large, the number of samples becomes nearly uniform across different vehicle counts, but intuitively, vehicles that are too far away are unlikely to interact meaningfully with pedestrians. Therefore, to maximize the amount of realistic interaction data while keeping the experimental setup manageable, we select scenes with 1–4 vehicles within a 0–15 m range for %both one-person and two-person scenes
one-, two-, and three-person scenes, %\todo{for 3-ped as well?} 
resulting in a total of $3\times4 $ experimental conditions. \GX{This choice reflects a practical trade-off between training set size and empirically observed performance, which we will further justify with evidence from prior studies in future work.}The number of segmented scenes for each condition is summarized in Table~\ref{tab:num_samples}.

In summary, the processed dataset for training and validation includes all possible pedestrian combinations together with the actual number of vehicles within a fixed range. %Although the data are insufficient to capture the effects of pedestrian–pedestrian interactions, 
Although the proposed scene segmentation approach may limit the potential for capturing pedestrian–pedestrian interactions, the inclusion of real vehicle information makes it well-suited for evaluating pedestrian–vehicle interactions.

\section{Method}
\begin{figure*}[!htb]
  \centering
  \includegraphics[width=\textwidth]{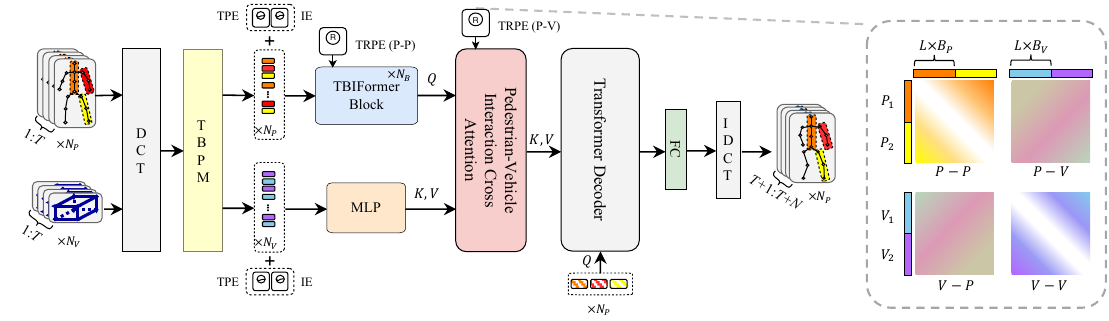}
  \caption{\textbf{Overview of the proposed Vehicle-conditioned pedestrian pose forecasting network.} The model receives 3D pedestrian poses and 3D vehicle bounding box data, transforms them into displacement sequences, and applies Discrete Cosine Transform (DCT) to discard high-frequency components for a more compact representation. Pedestrian and vehicle features are then processed through separate encoders and fused via cross-attention before being decoded into future pedestrian poses.}
  \label{fig:network}
\end{figure*}

In this section, we will introduce our proposed vehicle-conditioned pedestrian pose forecasting network as illustrated in Fig.~\ref{fig:network}. %We first define the problem in Section.... then, ... 
We first define the problem in Section~\ref{sec:problem} then present the overall framework in Section~\ref{sec:framework}, and further describe its core components in the following subsections.

\subsection{Problem Definition}
\label{sec:problem}

Consider $P$ pedestrians with observed 3D skeletal poses over $T+1$ frames, denoted as $X_p^{1:T+1} = \{x_p^1, x_p^2, \dots, x_p^{T+1}\}$ for $p = 1, \dots, P$. To capture motion dynamics, we represent pedestrian motion as frame-to-frame displacements $y^i = x^{i+1} - x^i$, forming the displacement sequence $Y^{1:T} = \{y^1, \dots, y^T\}$. Similarly, for $K$ surrounding vehicles, we observe their 3D trajectories $V^{1:T+1} = \{v_1^{1:T+1}, \dots, v_K^{1:T+1}\}$ in the same coordinate space, where each $v_k^t$ contains the 8 corner points of the vehicle's 3D bounding box at frame $t$, and convert them to displacement sequences $U^{1:T} = \{u^1, \dots, u^T\}$ with $u^i_k = v_k^{i+1} - v_k^i$. The goal is to leverage both pedestrian and vehicle displacement sequences, $(Y^{1:T}, U^{1:T})$, to predict the $N$ future pedestrian displacements $Y^{T+1:T+N}$, which are then transformed back into 3D poses $X^{T+2:T+N+1}$.

\subsection{Overall Framework}
\label{sec:framework}

%As illustrated in Fig.~\ref{fig:network}, we extend the original TBIFormer~\cite{peng2023trajectory} to incorporate vehicle information for pedestrian 3D pose forecasting. 
%\eh{any justification of using TBIFormer?} 
As illustrated in Fig.~\ref{fig:network}, we extend TBIFormer\cite{peng2023trajectory}, which effectively captures long-term temporal dependencies and fine-grained body-part dynamics, to incorporate vehicle information for pedestrian 3D pose forecasting.
For the pedestrian branch, our network retains the core TBIFormer architecture, including stacked TBIFormer blocks and the Temporal Body Partition Module (TBPM)%, and the Transformer decoder
. To leverage vehicle information, we introduce an additional \textbf{vehicle branch} that processes displacement sequences of vehicles in the same 3D coordinate space as pedestrians. Pedestrian and vehicle features are then fused through a \textbf{Pedestrian–Vehicle Interaction Cross-Attention (PVI-CA)} module, which models interactions between pedestrian body parts and vehicle groups. Notably, we extend the Trajectory-Aware Relative Position Encoding (TRPE) to the pedestrian–vehicle cross attention, providing discriminative spatial and temporal cues for accurate pose prediction. The fused representation is subsequently fed into the traditional Transformer decoder~\cite{vaswani2017attention} to predict future pedestrian poses. Overall, our model explicitly integrates vehicle motion while preserving the temporal, spatial, and social modeling capabilities of TBIFormer.

\subsection{Vehicle Encoder}
To incorporate vehicle information into the same 3D space as pedestrians, we design a %simple 
Vehicle Encoder that converts 3D vehicle bounding box data into a compact feature representation suitable for interaction modeling. Each vehicle is represented by the 8 corner points of its 3D bounding box. These corner points are first used to compute a displacement sequence, capturing the vehicle’s temporal motion across consecutive frames. The displacement sequence is then transformed via Discrete Cosine Transformation (DCT)~\cite{ahmed2006discrete}, which suppresses high-frequency components and yields a more compact representation in the displacement trajectory space, preserving the primary motion trends while reducing noise and redundancy. The resulting sequence is downsampled along the temporal dimension to length $L$. Following this, the 8 corner points are divided into $B_V=12$ logical groups (12 edges of the bounding box)
% \todo{pls update this part and justify the choice of the veh representation by referring the read to the ablation study}
%\eh{why 2 groups? how those 8 points are divided into the 2 groups?}\todo{Actually,the 8 corner points could be divided into 1, 2, 4, or 8 groups. Using 1 group is similar to just using the center, while 4 groups would produce wrong outputs because, with the pooling parameters kept the same as in the pedestrian branch (kernel size = 5, padding = 1), each group would contain 2 points, resulting in zero features. By the way, I think we can omit further details here, just explain how.} \eh{for 1, you can refer to ablation to say it is not optimal; for 2, again, ablation to say it is the best; }
, a choice motivated by the ablation study in Section~\ref{sec:ablation} and analogous to the body-part division (Temporal Body Partition Module) used for pedestrians as in TBIFormer (i.e. 3D keypoints of each pedestrian are divided into 5 groups to represent the trunk and 4 limbs), allowing the network to model localized motion patterns within the vehicle structure. For $N_V$ vehicles in the scene, concatenating all groups across all vehicles forms a Multi-Vehicle Feature sequence of length $U = N_v \times B_v \times L$ with feature dimension $D$. Temporal positional encoding (TPE) and identity encoding (IE) are applied to each sequence in the same manner as the pedestrian branch, ensuring temporal dynamics and individual vehicle identity are preserved. The Multi-Vehicle Feature sequence is then processed through a two-layer MLP to extract high-dimensional embeddings, ensuring the same feature dimension as the pedestrian branch, which facilitates seamless integration in the subsequent cross-attention module, enabling effective fusion of pedestrian and vehicle information for interaction-aware 3D pose forecasting.
%To incorporate vehicle information into the same 3D space as pedestrians, we design a simple Vehicle Encoder. 
% Each vehicle is represented by its 8 corner points of 3D box, which are first processed to obtain a displacement sequence and then transformed via Discrete Cosine Transformation (DCT)~\cite{ahmed2006discrete}, discarding high-frequency information for a more compact representation in the displacement trajectory space. The 8 points are then split into two groups, analogous to body-part division in pedestrians, and processed through a two-layer MLP. Before the MLP, we apply temporal positional encoding(TPE) and identity encoding(IE) in the same manner as the pedestrian branch. The vehicle features are kept at the same feature dimension as the pedestrian branch, allowing later fusion in the cross-attention module.

\subsection{Pedestrian–Vehicle Interaction Cross-Attention}

To effectively incorporate vehicle information into pedestrian motion forecasting, we extend the original Trajectory-Aware Relative Position Encoding (TRPE) to model interactions between pedestrians and vehicles. The extended TRPE encodes trajectory-aware relational information between each pedestrian and surrounding vehicles, capturing both spatial and temporal context, as illustrated in Fig.~\ref{fig:network}.

The resulting trajectory-aware relative position embeddings are then integrated into a cross-attention mechanism, referred to as Pedestrian-Vehicle Interaction Cross-Attention (PVI-CA). Given the pedestrian features $H^p$ from the pedestrian branch and vehicle features $H^v$ from the vehicle branch, the cross-attention computes:
\begin{equation}
    Q = H^p W_Q, \quad K = H^v W_K, \quad V = H^v W_V,
\end{equation}
\begin{equation}
    \text{PVI-CA}(Q,K,V) = \text{softmax}\Big(\frac{Q K^\top + B_{\text{TRPE}}}{\sqrt{d_z}}\Big) V,
\end{equation}
where $B_{\text{TRPE}}$ is the contextual bias derived from the extended TRPE between pedestrian and vehicle features (see Fig.~\ref{fig:network}). By integrating both spatial and trajectory-aware relational information, PVI-CA allows the model to selectively attend to vehicles that are most relevant to each pedestrian's motion.

\subsection{Decoder}
As illustrated in Fig.~\ref{fig:network}, we follow the standard Transformer decoder design~\cite{vaswani2017attention}. 
Specifically, the joint coordinates of the last observed pedestrian sub-sequence are concatenated and down-sampled by a 1D convolution to form global body query tokens, while the fused pedestrian–vehicle features from the PVI-CA module serve as keys and values. The decoder encodes the relations between the current queries and historical context, conditioned on pedestrian and vehicle information. Finally, two fully connected layers followed by an Inverse Discrete Cosine Transformation (IDCT)~\cite{ahmed2006discrete} generate the future motion trajectory $X_{T+2:T+N+1}$ for each pedestrian.

\subsection{Training Objective}
\label{sec:loss}

We treat the vehicle motion as a conditioning signal and therefore do not predict its future trajectory. As a result, the training objective is applied only to pedestrian poses.

We optimize the model using a reconstruction loss based on the Mean Per Joint Position Error (MPJPE). For a single training sample, the loss is defined as
\begin{equation}
\mathcal{L}_{\mathrm{rec}}
=\frac{1}{J\,*N}\sum_{t=T+1}^{T+N}\sum_{j=1}^{J}
\left\|\hat{\mathbf{y}}_{t,j}-\mathbf{y}_{t,j}\right\|_2,
\end{equation}

where $\hat{\mathbf{y}}_{t,j}$ and $\mathbf{y}_{t,j}$ denote the estimated and ground-truth pose displacements of joint $j$ at time step $t$, respectively, and $J$ is the number of body joints.

\section{Experiment}
\subsection{Implementation Details}

We implement our framework in PyTorch, and the experiments are performed on a single Nvidia GeForce RTX 4090 GPU. We train our model for 50 epochs using the Adam optimizer with a batch size of 32, a learning rate of $2 \times 10^{-5}$, and a dropout rate of 0.2. All other settings remain consistent with TBIFormer~\cite{peng2023trajectory}, with the model trained for 2\,s (50 frames) and evaluated for 1\,s (25 frames) prediction.

\subsection{Baselines}
Since no prior work directly addresses 3D pose forecasting with vehicle context, we adapt the state-of-the-art multi-agent 3D pose forecasting method TBIFormer~\cite{peng2023trajectory} as our baseline. Both TBIFormer and our model are trained and evaluated on the one-, two-, and three-pedestrian scenarios extracted from Waymo-3DSkelMo and the Waymo Dataset, following the dataset procedure described in Section~\ref{sec:scene_seg}. In each scenario, TBIFormer is applied in the pedestrian-only setting, while our model uses pedestrian + vehicle(s). For both methods, we use a 2\,s input and predict 1\,s ahead, with training and evaluation conducted under varying numbers of vehicles.

\subsection{Metrics}

We evaluate our predictions using three widely used metrics, which capture different aspects of pose and trajectory accuracy.

\textbf{JPE (Joint Position Error):} Evaluates both global and local pose predictions by averaging the $L_2$ distance of all joints at each predicted timestep:
\begin{equation}
\text{JPE} = \frac{1}{T \cdot N_j} \sum_{t=1}^{T} \sum_{j=1}^{N_j} \| \hat{\mathbf{p}}_{t}^{(j)} - \mathbf{p}_{t}^{(j)} \|_2
\end{equation}
where $T$ is the number of predicted timesteps, $N_j$ is the number of joints, $\hat{\mathbf{p}}_{t}^{(j)}$ and $\mathbf{p}_{t}^{(j)}$ are the predicted and ground-truth positions of joint $j$ at timestep $t$, respectively.

\textbf{APE (Aligned Pose Error):} Evaluates the forecasted local motion by measuring the average $L_2$ distance of all joints, after removing global translation to capture pure pose error:
\begin{equation}
\text{APE} = \frac{1}{T \cdot N_j} \sum_{t=1}^{T} \sum_{j=1}^{N_j} \| (\hat{\mathbf{p}}_{t}^{(j)} - \hat{\mathbf{r}}_t) - (\mathbf{p}_{t}^{(j)} - \mathbf{r}_t) \|_2
\end{equation}
where $\hat{\mathbf{r}}_t$ and $\mathbf{r}_t$ are the predicted and ground-truth root positions at timestep $t$.

\textbf{FDE (Final Displacement Error):} Measures the accuracy of the forecasted global trajectory by computing the $L_2$ distance of the root position at the final predicted timestep.  
\begin{equation}
\text{FDE} = \| \hat{\mathbf{r}}_T - \mathbf{r}_T \|_2
\end{equation}

\subsection{Results}

\begin{table*}[htbp]
\centering
\caption{Evaluation metrics (in millimeters) for different prediction horizons under varying numbers of vehicles. Results compare one-, two-, and three-pedestrian scenarios, showing performance with and without surrounding vehicle information. Metrics include MPJPE, APE, and FDE %, averaged over the prediction frames at 0.2s, 0.6s, and 1.0s
at the prediction frames of 0.2s, 0.6s, and 1.0s.  Bold face indicates best performance.}
\label{tab:combined_metrics}
\resizebox{\textwidth}{!}{
\begin{tabular}{@{}c|l||c c c |c c c ||c c c |c c c||c c c |c c c@{}}
\toprule
\multirow{3}{*}{\textbf{\thead{\#\\of\\Veh.}}} & \multirow{3}{*}{\textbf{Metric}} 
 & \multicolumn{6}{c||}{\textbf{1 Pedestrian Scene}} 
 & \multicolumn{6}{c||}{\textbf{2 Pedestrians Scene}}
 & \multicolumn{6}{c}{\textbf{3 Pedestrians Scene}} 
 \\
\cmidrule(lr){3-8} \cmidrule(lr){9-14} \cmidrule(lr){15-20}
 & 
 & \multicolumn{3}{c|}{\textbf{~\cite{peng2023trajectory} (Ped. only)}} 
 & \multicolumn{3}{c||}{\textbf{Ours}} 
  & \multicolumn{3}{c|}{\textbf{~\cite{peng2023trajectory} (Ped. only)}} 
 & \multicolumn{3}{c||}{\textbf{Ours }} 
 & \multicolumn{3}{c|}{\textbf{~\cite{peng2023trajectory} (Ped. only)}} 
 & \multicolumn{3}{c}{\textbf{Ours }} \\
\cmidrule(lr){3-5} \cmidrule(lr){6-8} \cmidrule(lr){9-11} \cmidrule(lr){12-14} \cmidrule(lr){15-17} \cmidrule(lr){18-20}
 % & & 0.2s & 0.6s & 1s & Overall 
   && 0.2s & 0.6s & 1.0s 
  & 0.2s & 0.6s & 1.0s 
  & 0.2s & 0.6s & 1.0s 
   & 0.2s & 0.6s & 1.0s 
  & 0.2s & 0.6s & 1.0s 
  & 0.2s & 0.6s & 1.0s \\

   % & 0.2s & 0.6s & 1s & Overall 
   % & 0.2s & 0.6s & 1s & Overall 
   % & 0.2s & 0.6s & 1s & Overall \\
\midrule
\multirow{3}{*}{\textbf{1}} 
  % & MPJPE & 85 & 225 & 306 & 205 & 75 & 216 & 295 & 195 & 88 & 228 & 324 & 213 & 79 & 215 & 304 & 199 \\
  % & APE   & 53 & 110 & 116 & 93  & 49 & 107 & 114 & 90  & 54 & 109 & 116 & 93  & 51 & 107 & 114 & 91 \\
  % & FDE   & 60 & 176 & 264 & 167 & 51 & 168 & 253 & 157 & 63 & 181 & 283 & 176 & 54 & 168 & 263 & 162 \\
    & MPJPE$\downarrow$ & 84 & 232 & 316  & \hi{78} & \hi{225} & \hi{311}  & 88 & 228 & 324  & \hi{79} & \hi{216} & \hi{304}      & 97 & 251 & 361  & \hi{82} & \hi{215} & \hi{299} \\
  & APE$\downarrow$   & 53 & 109 & 118   & \hi{49} & \hi{107} & \hi{115}   & 54 & 110 & 116   & \hi{52} & \hi{107} & \hi{115}    & 57 & 112 & \hi{119}  & \hi{55} & \hi{111} & \hi{119} \\
  & FDE$\downarrow$   & 57 & 183 & 273  & \hi{54} & \hi{179} & \hi{271}  & 63 & 181 & 283  & \hi{55} & \hi{170} & \hi{265}       & 72 & 206 & 323  & \hi{55} & \hi{165} & \hi{255} \\
\midrule
\multirow{3}{*}{\textbf{2}} 
  % & MPJPE & 91 & 235 & 320 & 215 & 81 & 225 & 305 & 204 & 89 & 207 & 275 & 190 & 76 & 203 & 273 & 184 \\
  % & APE   & 55 & 118 & 119 & 97  & 52 & 110 & 116 & 93  & 62 & 116 & 123 & 100 & 52 & 111 & 118 & 94 \\
  % & FDE   & 63 & 232 & 276 & 190 & 55 & 176 & 261 & 164 & 57 & 148 & 225 & 143 & 50 & 151 & 226 & 142 \\
    & MPJPE$\downarrow$ & 88 & 227 & \hi{302} & \hi{82} & \hi{224} & 303  & 86 & 225 & 313  & \hi{78} & \hi{205} & \hi{275}    & 94 & 238 & 334   & \hi{80} & \hi{208} & \hi{283} \\
  & APE$\downarrow$   & 56 & 112 & 118   & \hi{53} & \hi{111} & \hi{117}   & 54 & 112 & 120  & \hi{53} & \hi{110} & \hi{117}     & 59 & 118 & 127  & \hi{55} & \hi{114} & \hi{123} \\
  & FDE$\downarrow$   & 61 & 178 & \hi{259}  & \hi{56} & \hi{175} & \hi{259}  & 60 & 175 & 273  & \hi{52} & \hi{152} & \hi{229}     & 66 & 187 & 290  & \hi{53} & \hi{155} & \hi{236} \\
\midrule
\multirow{3}{*}{\textbf{3}} 
  % & MPJPE & 86 & 252 & 292 & 210 & 80 & 214 & 293 & 196 & 89 & 230 & 315 & 211 & 76 & 208 & 285 & 190 \\
  % & APE   & 55 & 123 & 126 & 101 & 52 & 113 & 121 & 95 & 57 & 115 & 123 & 98 & 52 & 113 & 123 & 96 \\
  % & FDE   & 58 & 202 & 245 & 168 & 54 & 162 & 247 & 154 & 61 & 180 & 270 & 170 & 51 & 156 & 237 & 148 \\
    & MPJPE$\downarrow$ & 87 & 220 & 300  & \hi{79} & \hi{213} & \hi{293}  & 89 & 231 & 323  & \hi{76} & \hi{208} & \hi{285}    & 93 & 237 & 334  & \hi{79} & \hi{201} & \hi{272} \\
  & APE$\downarrow$   & 54 & 114 & 123  & \hi{52} & \hi{113} & \hi{121}  & 56 & 115 & 124 & \hi{52} & \hi{112} & \hi{122}     & 57 & 117 & \hi{125}  & \hi{56} & \hi{115} & \hi{125} \\
  & FDE$\downarrow$   & 61 & 168 & 254  & \hi{53} & \hi{161} & \hi{248}  & 61 & 178 & 276  & \hi{50} & \hi{155} & \hi{236}    & 65 & 184 & 284  & \hi{50} & \hi{147} & \hi{221} \\
\midrule
\multirow{3}{*}{\textbf{4}} 
  % & MPJPE & 84 & 245 & 280 & 203 & 76 & 204 & 283 & 188 & 86 & 215 & 293 & 198 & 76 & 204 & 281 & 187 \\
  % & APE   & 55 & 119 & 121 & 98  & 51 & 110 & 118 & 93  & 55 & 111 & 122 & 96 & 51 & 109 & 119 & 93 \\
  % & FDE   & 56 & 192 & 231 & 160 & 50 & 155 & 238 & 148 & 57 & 163 & 244 & 155 & 51 & 154 & 236 & 147 \\

& MPJPE$\downarrow$ & 83 & 206 & \hi{277}  & \hi{77} & \hi{205} & 285  & 92 & 242 & 348  & \hi{76} & \hi{202} & \hi{278}    & 99 & 253 & 364  & \hi{78} & \hi{207} & \hi{279} \\
  & APE$\downarrow$   & 54 & 112 & 121  & \hi{52} & \hi{110} & \hi{119}   & 55 & 113 & 122  & \hi{51} & \hi{109} & \hi{119}    & 58 & 119 & 131  & \hi{54} & \hi{115} & \hi{125} \\
  & FDE$\downarrow$   & 55 & \hi{152} & \hi{229}  & \hi{51} & 155 & 241  & 67 & 197 & 307  & \hi{50} & \hi{151} & \hi{230}    & 71 & 203 & 316  & \hi{52} & \hi{152} & \hi{229} \\
\bottomrule
\end{tabular}
}
\end{table*}

%We conduct experiments separately for one-pedestrian and two-pedestrian scenarios
We conduct experiments separately for one-, two-, and three-pedestrian scenarios, comparing prediction performance with and without surrounding vehicles under varying numbers of vehicles. The evaluation metrics at different prediction horizons are summarized in Table~\ref{tab:combined_metrics}. From the results, several observations can be made.

Firstly, the inclusion of vehicles consistently improves overall prediction accuracy across all metrics. %For example, in the one-pedestrian scenario with a single vehicle, the overall MPJPE decreases from 205\,mm (pedestrian only) to 195\,mm (pedestrian + vehicle), representing an improvement of approximately 4.9\%. 
For example, in the one-pedestrian scenario with a single vehicle, the MPJPE decreases from 84–316\,mm (pedestrian only) to 78–311\,mm (ours), corresponding to an improvement of around 1.5\%–7\%. Similar improvements are observed for APE and FDE, with reductions of around 0.7\%–7.5\% overall. This trend holds across different numbers of vehicles and all pedestrian scenarios, with overall improvements ranging from 1.5\%-21\% for MPJPE, 1.8\%-12\% for APE, and 0.7\%-15\% for FDE, depending on the number of vehicles in the scenes. However, in the one-pedestrian scenario with four vehicles, the prediction performance at 1.0s slightly decreases. \GX{Since the four-vehicle outperforms the two- and three-pedestrian scenarios, this is unlikely due to limited capacity for multiple contextual agents and is more likely because some vehicles are distant or less relevant, which introduces noise to the training scenarios.}
% likely because some vehicles are distant or less relevant, which introduces noise to the training data.

Secondly, %comparing the one- and two-pedestrian scenarios, the overall improvements are generally slightly higher in the one-pedestrian case. 
comparing the one-, two-, and three-pedestrian scenarios, the overall improvements generally increase with the number of pedestrians. %For example, in the one-pedestrian case, MPJPE at 1.0s decreases from 316 mm (pedestrian only) to 311 mm (ours), corresponding to an improvement of about 1.6\%. In the two-pedestrian case, MPJPE decreases from 313 mm to 275 mm ($\sim 12\%$ improvement), and in the three-pedestrian case, it decreases from 334 mm to 283 mm ($\sim 15\%$ improvement). 
For example, in the one-pedestrian scenario with a single vehicle, the average MPJPE across 0.2s, 0.6s, and 1.0s decreases from 211 mm (pedestrian only) to 205 mm (ours), corresponding to an improvement of about 2.9\%. In the two-pedestrian scenario with one vehicle, the average MPJPE decreases from 213 mm to 200 mm ($\sim 6.4\%$ improvement), and in the three-pedestrian scenario with one vehicle, it decreases from 236 mm to 199 mm ($\sim 15.9\%$ improvement).
This trend is likely because multi-pedestrian scenes involve more complex interactions, allowing vehicle information to provide stronger contextual cues, whereas in single-pedestrian scenes, the interaction context from vehicles is more limited.

% This difference is likely due to variations in vehicle coverage and distribution around the selected pedestrian subsets. One-pedestrian scenes tend to include more samples with pedestrians closer to vehicles, allowing vehicle information to have a more pronounced impact. In contrast, two-pedestrian scenes include relatively fewer samples where both pedestrians are near vehicles simultaneously, leading to slightly smaller overall improvements. 
% \eh{are there any statistics on the avg distance between pedestrians and vehicles to support this analysis?}
% \todo{No statistics here, but we can find this situation from our sampling strategy, because the choice of vehicle is based on the distance from any one person, so there must be a vehicle that is far away from another person.}

Thirdly, we examine the effect of increasing vehicle count within each scenario by focusing on relative improvements, as absolute metrics are not directly comparable due to differences in the underlying training data for each vehicle count. We observe that, overall, relative improvements increase with the number of vehicles. %For instance, in the one-pedestrian scenario, including a second vehicle increases the overall MPJPE improvement from 4.9\% (one vehicle) to 5.8\% (two vehicles). 
For instance, in the two-pedestrian scenario, increasing the number of surrounding vehicles leads to steadily larger improvements in overall MPJPE, with the improvement growing from approximately 6\% with one vehicle to around 18\% with four vehicles. A similar trend is observed in the three-pedestrian scenario, where additional vehicles also result in substantial performance gains. In contrast, in the one-pedestrian scenario, the improvements are relatively small and even show slight decreases when adding more than three vehicles, indicating that the contribution of vehicle information is limited when only a single pedestrian is present.
% For instance, in the two-pedestrian scenario, including a second vehicle increases the overall MPJPE improvement from approximately 6\% with one vehicle to about 12\% with two vehicles.
% However, adding a third or fourth vehicle does not lead to further improvement and can even slightly reduce performance, highlighting the importance of focusing on vehicles that are more directly interacting with the pedestrians rather than simply increasing vehicle count.

Overall, the results confirm that vehicle information plays a positive role in improving pedestrian trajectory prediction. Compared with the baseline, our model more effectively leverages such contextual cues, leading to consistent performance gains.

\begin{table}[ht]
\centering
\caption{Ablation studies on different components of Our model. Our full method and its variants are evaluated on the 2 pedestrians and 3 vehicles subdataset (averaged over 0.2s, 0.4s, 0.6s, 0.8s, 1.0s). Bold face indicates best performance.}
\label{tab:ablation}

\begin{tabular}{lccc}
\toprule
\textbf{Method} & MPJPE$\downarrow$ & APE$\downarrow$ & FDE$\downarrow$ \\
\midrule
% TBIFormer               & 198 & 96  & 155 \\
% + Vehicle Center        & 231 & 98  & 192 \\
TBIFormer               & 218.8 & 103.0  & 173.4 \\
 + Vehicle Center        & 224.4 & 105.2 & 177.8 \\

\midrule
% + Vehicle 8 pts         & 218 & 106 & 172 \\
+ Veh Branch w/o TRPE   & 195.4 & 100.4  & 150.2 \\
\midrule

+ Veh Branch w TRPE  \\
\hspace{1em}1 group   &195.4 &100.2 &150.0 \\
\hspace{1em}2 group   &194.8 &100.2 &150.6 \\
\hspace{1em}4 group   &195.4 &100.2 &150.0 \\
\hspace{1em}6 group   &195.6 &100.4 &150.6 \\
\hspace{1em}8 group   &195.4 &100.0 &150.2 \\
\hspace{1em}12 group  &194.8 &100.0 &149.8 \\
\midrule
Full Model (with TRPE) & \textbf{194.8} & \textbf{100.0}  & \textbf{149.8} \\
\bottomrule
\end{tabular}
\end{table}

\subsection{Ablation Studies}
\label{sec:ablation}

Table~\ref{tab:ablation} shows ablation studies on the two-pedestrian, three-vehicle subset.  
% Simply treating the vehicle center as a pedestrian body part and using the original TBIFormer for self-attention increases errors (MPJPE 231 vs. 198, APE 98 vs. 96, FDE 192 vs. 155). Replacing it with full 8-point vehicle representations improves performance compared to using only the center (MPJPE 218, APE 106, FDE 172), indicating that richer vehicle geometry provides more useful cues. Introducing a dedicated vehicle branch without PVI-CA, using only conventional cross-attention, further reduces errors (MPJPE 188, APE 93, FDE 147), demonstrating that explicit vehicle modeling benefits pedestrian–vehicle interaction learning. Adding PVI-CA yields marginal improvement in MPJPE (187 vs. 188) while keeping APE and FDE unchanged, suggesting that the proposed temporal-vehicle relational encoding provides limited but consistent benefits.
Simply treating the vehicle center as a pseudo pedestrian body part and applying the original TBIFormer~\cite{peng2023trajectory} increases MPJPE, APE, and FDE by roughly 2.5\% each, indicating that naively treating the vehicle as a pedestrian provides insufficient cues for accurate pedestrian–vehicle interaction modeling.
% Simply treating the vehicle center as a pseudo pedestrian body part and applying the original TBIFormer increases MPJPE by nearly $9\% $ and FDE by approximately $12\% $, while APE shows little change, indicating that naively treating the vehicle as a pedestrian provides insufficient cues for accurate pedestrian–vehicle interaction modeling.

Introducing a dedicated vehicle branch with conventional cross-attention but without TRPE in it substantially reduces errors, with MPJPE and FDE decreasing by about $10.7\% $ and $13.4\% $, respectively, demonstrating that explicit vehicle modeling effectively improves performance.
Adding TRPE within the cross-attention, thereby forming PVI-CA, brings marginal but consistent improvements under different grouping configurations of the vehicle bounding box corners. Here, \textit{1 group} treats all 8 corners as a single group; \textit{2 groups} split the corners into front and back faces; \textit{4 groups} represent the four vertical edges of the bounding box; \textit{6 groups} treat each face of the bounding box as a group; \textit{8 groups} treat each corner individually; and \textit{12 groups} consider each bounding box edge as a separate group. It can be seen that the performance differences across different groupings are small, and our full model adopts the \textit{12 groups} setting, \GX{as it best captures the spatial geometry and scale of surrounding vehicles without much extra computation,} though all group configurations are supported.
% \GX{ I will talk more and explain how the group is divided}

% Table~\ref{tab:ablation} shows ablation studies on the two-pedestrian, four-vehicle subset. Simply adding vehicle center increases errors, while using full 8-point vehicle representations improves performance, indicating that richer vehicle geometry provides more useful cues. Introducing a dedicated vehicle branch without TRPE further reduces MPJPE, APE, and FDE, demonstrating that explicit vehicle modeling benefits pedestrian--vehicle interaction learning. Adding TRPE yields marginal improvement in MPJPE, suggesting that temporal relational encoding provides limited but consistent benefits.

\begin{figure*} [htb]
    \centering
    \includegraphics[width=\textwidth]{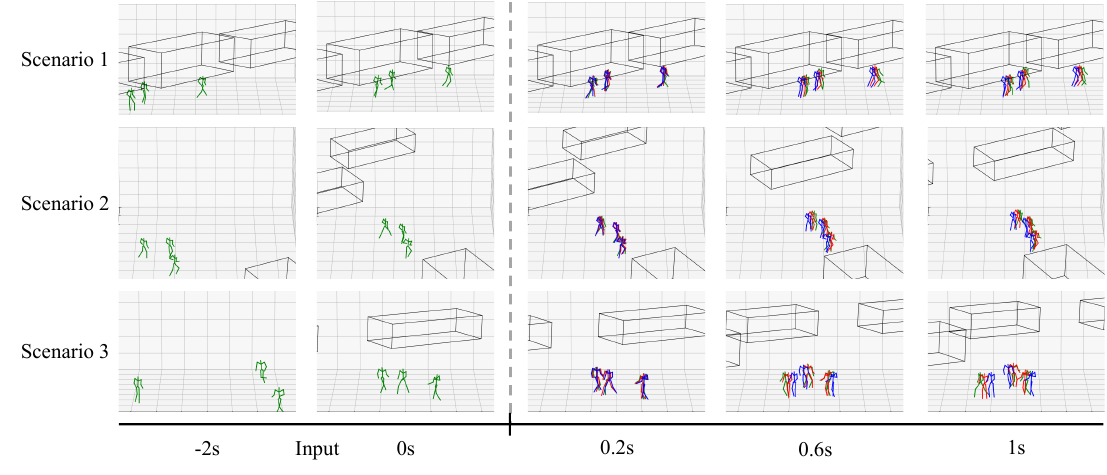}
    % \caption{Qualitative comparison between TBIFormer and our model. Each pedestrian is shown with a distinct color pair: observed history in a blue-family shade, predicted future in corresponding red shades, and ground truth in green. Vehicles are shown as boxes. Trajectories are sampled every 10 frames.}
    % \caption{Qualitative comparison between TBIFormer~\cite{peng2023trajectory} and our model. Each pedestrian is shown in a distinct color. In the ground truth and our model, black boxes in the background represent surrounding vehicles used as input to our model, whereas TBIFormer does not use vehicle information. \GX{Orange boxes highlight cases with large discrepancies between TBIFormer's predictions and the ground truth.}}
    \caption{Qualitative comparison between TBIFormer~\cite{peng2023trajectory} and our model. Ground truth trajectories are shown in green, TBIFormer predictions in blue, and ours in red. Black boxes denote surrounding vehicles used by our model as input, whereas TBIFormer does not use vehicle information.}
    \label{fig:vis}
\end{figure*}

\subsection{Visualization}
We visualize one scenario involving three pedestrians and four vehicles, and compare the predictions of TBIFormer~\cite{peng2023trajectory} without vehicles and our model with vehicles. From the qualitative results in Fig.~\ref{fig:vis}, we observe that both TBIFormer and our model perform well for short-duration predictions (0.2s), while our model shows clear advantages for longer-duration predictions (0.6s and 1s). 
% TBIFormer’s predictions tend to lead the ground truth by a noticeable margin (orange boxes in Fig.~\ref{fig:vis}), whereas our predictions more accurately follow the true trajectories. In particular, in the 1.0s prediction, TBIFormer without vehicle information predicts motion trends that move away from pedestrians while approaching nearby vehicles, which could raise safety concerns. In contrast, our model better respects the surrounding vehicle positions, highlighting the importance of incorporating vehicle information and the effectiveness of explicitly modeling pedestrian–vehicle interactions.
TBIFormer’s predictions tend to lead the ground truth by a noticeable margin, whereas our predictions more accurately follow the true trajectories. In particular, in the 1.0s prediction, TBIFormer predicts more conservative, delayed motion trends. In contrast, our model better respects the positions of surrounding vehicles and predicts more accurate motions that align more closely with the ground truth, highlighting the importance of incorporating vehicle information and explicitly modeling pedestrian–vehicle interactions.

\section{Conclusion}

% A conclusion section is not required. Although a conclusion may review the main points of the paper, do not replicate the abstract as the conclusion. A conclusion might elaborate on the importance of the work or suggest applications and extensions. 
In this work, we address the critical problem of 3D pedestrian pose forecasting in autonomous driving scenarios, emphasizing the often-overlooked influence of interactions between 3D pedestrian poses and 3D vehicles on pedestrian behavior. We enhance the Waymo-3DSkelMo dataset by incorporating 3D vehicle bounding boxes and a sampling strategy, enabling realistic modeling of pedestrian–vehicle interactions. Building on the TBIFormer architecture, we propose a Vehicle-conditioned 3D pose forecasting network, where pedestrian predictions are conditioned on not only their historical motion but also the surrounding vehicle context. Extensive experiments demonstrate that incorporating vehicle information significantly improves the accuracy of predicted pedestrian poses and validates different approaches for modeling pedestrian–vehicle interactions. 
\GX{Although we focus exclusively on vehicles and omit other contextual cues, such as static scene structure and other road users (e.g., cyclists), our findings indicate that vehicle context provides impactful information beyond motion history alone, and we believe integrating richer scene context is a promising direction for future work.}
Our work highlights the importance of multi-agent context, particularly vehicles, in accurate pedestrian motion prediction and provides a foundation for safer autonomous driving systems.


%%%%%%%%%%%%%%%%%%%%%%%%%%%%%%%%%%%%%%%%%%%%%%%%%%%%%%%%%%%%%%%%%%%%%%%%%%%%%%%%

\section*{Acknowledgments}
Guangxun Zhu is supported by the funding from the China Scholarship Council (CSC).

%Bibliography
\bibliographystyle{unsrt}  
\bibliography{references}

@INPROCEEDINGS{SocialCircle,
  author={Wong, Conghao and Xia, Beihao and Zou, Ziqian and Wang, Yulong and You, Xinge},
  booktitle={2024 IEEE/CVF Conference on Computer Vision and Pattern Recognition (CVPR)}, 
  title={SocialCircle: Learning the Angle-based Social Interaction Representation for Pedestrian Trajectory Prediction}, 
  year={2024},
  volume={},
  number={},
  pages={19005-19015},
  doi={10.1109/CVPR52733.2024.01798}}

@article{azarmi2024pip,
  author={Azarmi, Mohsen and Rezaei, Mahdi and Wang, He},
  journal={IEEE Transactions on Intelligent Transportation Systems}, 
  title={PIP-Net: Pedestrian Intention Prediction in the Wild}, 
  year={2025},
  volume={26},
  number={7},
  pages={9824-9837},
  doi={10.1109/TITS.2025.3570794}
}

@ARTICLE{Crosato:TIV2023,
  author={Crosato, Luca and Shum, Hubert P. H. and Ho, Edmond S. L. and Wei, Chongfeng},
  journal={IEEE Transactions on Intelligent Vehicles}, 
  title={Interaction-Aware Decision-Making for Automated Vehicles Using Social Value Orientation}, 
  year={2023},
  volume={8},
  number={2},
  pages={1339-1349},
  doi={10.1109/TIV.2022.3189836}}

@INPROCEEDINGS{PedFormer,
  author={Rasouli, Amir and Kotseruba, Iuliia},
  booktitle={2023 IEEE International Conference on Robotics and Automation (ICRA)}, 
  title={PedFormer: Pedestrian Behavior Prediction via Cross-Modal Attention Modulation and Gated Multitask Learning}, 
  year={2023},
  volume={},
  number={},
  pages={9844-9851},
  doi={10.1109/ICRA48891.2023.10161318}}

@INPROCEEDINGS{Distillation,
      title={Multi-modal Knowledge Distillation-based Human Trajectory Forecasting}, 
      author={Jaewoo Jeong and Seohee Lee and Daehee Park and Giwon Lee and Kuk-Jin Yoon},
      year={2025},
       booktitle={2025 IEEE/CVF Conference on Computer Vision and Pattern Recognition (CVPR)},  
}

@InProceedings{saadatnejad2023social,
      title={Social-Transmotion: Promptable Human Trajectory Prediction}, 
      author={Saadatnejad, Saeed and Gao, Yang and Messaoud, Kaouther and Alahi, Alexandre},
      booktitle={International Conference on Learning Representations (ICLR)},
      year={2024},
}

@INPROCEEDINGS{Jeong:CVPR2024,
  author={Jeong, Jaewoo and Park, Daehee and Yoon, Kuk-Jin},
  booktitle={2024 IEEE/CVF Conference on Computer Vision and Pattern Recognition (CVPR)}, 
  title={Multi-Agent Long-Term 3D Human Pose Forecasting via Interaction-Aware Trajectory Conditioning}, 
  year={2024},
  volume={},
  number={},
  pages={16975-16984},
  doi={10.1109/CVPR52733.2024.00160}}

@INPROCEEDINGS {MuPoTs-3D,
author = { Mehta, Dushyant and Sotnychenko, Oleksandr and Mueller, Franziska and Xu, Weipeng and Sridhar, Srinath and Pons-Moll, Gerard and Theobalt, Christian },
booktitle = { 2018 International Conference on 3D Vision (3DV) },
title = {{ Single-Shot Multi-person 3D Pose Estimation from Monocular RGB }},
year = {2018},
volume = {},
ISSN = {},
pages = {120-130},
doi = {10.1109/3DV.2018.00024},
url = {https://doi.ieeecomputersociety.org/10.1109/3DV.2018.00024},
publisher = {IEEE Computer Society},
address = {Los Alamitos, CA, USA},
month =sep}

@misc{mocapdata,
  title = {CMU Graphics Lab Motion Capture Database},
  author = {{CMU}},
  year = {2003},
  url = {http://mocap.cs.cmu.edu/}
}

@misc{waymo_open_dataset,
title = {Waymo Open Dataset: An autonomous driving dataset},
website = {\url{https://www.waymo.com/open}},
year = {2019}
}

@incollection{loper2023smpl,
  title={SMPL: A skinned multi-person linear model},
  author={Loper, Matthew and Mahmood, Naureen and Romero, Javier and Pons-Moll, Gerard and Black, Michael J},
  booktitle={Seminal Graphics Papers: Pushing the Boundaries, Volume 2},
  pages={851--866},
  year={2023}
}

@inproceedings{peng2023trajectory,
  title={Trajectory-aware body interaction transformer for multi-person pose forecasting},
  author={Peng, Xiaogang and Mao, Siyuan and Wu, Zizhao},
  booktitle={Proceedings of the IEEE/CVF conference on computer vision and pattern recognition},
  pages={17121--17130},
  year={2023}
}

@inproceedings{Waymo-3DSkelMo,
author = {Zhu, Guangxun and Fan, Shiyu and Dai, Hang and Ho, Edmond S. L.},
title = {Waymo-3DSkelMo: A Multi-Agent 3D Skeletal Motion Dataset for Pedestrian Interaction Modeling in Autonomous Driving},
year = {2025},
publisher = {Association for Computing Machinery},
address = {New York, NY, USA},
booktitle = {Proceedings of the 33rd ACM International Conference on Multimedia},
location = {Dublin, Ireland},
series = {MM '25}
}

@inproceedings{NeMF,
 author = {He, Chengan and Saito, Jun and Zachary, James and Rushmeier, Holly and Zhou, Yi},
 booktitle = {Advances in Neural Information Processing Systems},
 editor = {S. Koyejo and S. Mohamed and A. Agarwal and D. Belgrave and K. Cho and A. Oh},
 pages = {4244--4256},
 publisher = {Curran Associates, Inc.},
 title = {NeMF: Neural Motion Fields for Kinematic Animation},
 volume = {35},
 year = {2022}
}

@article{fan2023lidar,
    title={LiDAR-HMR: 3D Human Mesh Recovery from LiDAR},
    author={Fan, Bohao and Zheng, Wenzhao and Feng, Jianjiang and Zhou, Jie},
    journal={arXiv preprint arXiv:2311.11971},
    year={2023}
}

@inproceedings{Wang:NeurIPS2021,
author = {Wang, Jiashun and Xu, Huazhe and Narasimhan, Medhini and Wang, Xiaolong},
title = {Multi-person 3D motion prediction with multi-range transformers},
year = {2021},
isbn = {9781713845393},
publisher = {Curran Associates Inc.},
address = {Red Hook, NY, USA},
booktitle = {Proceedings of the 35th International Conference on Neural Information Processing Systems},
articleno = {462},
numpages = {14},
series = {NIPS '21}
}

@article{bentley1975multidimensional,
  title={Multidimensional binary search trees used for associative searching},
  author={Bentley, Jon Louis},
  journal={Communications of the ACM},
  volume={18},
  number={9},
  pages={509--517},
  year={1975},
  publisher={ACM New York, NY, USA}
}

@INPROCEEDINGS{UMPM,
  author={van der Aa, N.P. and Luo, X. and Giezeman, G.J. and Tan, R.T. and Veltkamp, R.C.},
  booktitle={2011 IEEE International Conference on Computer Vision Workshops (ICCV Workshops)}, 
  title={UMPM benchmark: A multi-person dataset with synchronized video and motion capture data for evaluation of articulated human motion and interaction}, 
  year={2011},
  volume={},
  number={},
  pages={1264-1269},
  doi={10.1109/ICCVW.2011.6130396}}

@article{ahmed2006discrete,
  title={Discrete cosine transform},
  author={Ahmed, Nasir and Natarajan, T\_ and Rao, Kamisetty R},
  journal={IEEE transactions on Computers},
  volume={100},
  number={1},
  pages={90--93},
  year={2006},
  publisher={IEEE}
}

@article{vaswani2017attention,
  title={Attention is all you need},
  author={Vaswani, Ashish and Shazeer, Noam and Parmar, Niki and Uszkoreit, Jakob and Jones, Llion and Gomez, Aidan N and Kaiser, {\L}ukasz and Polosukhin, Illia},
  journal={Advances in neural information processing systems},
  volume={30},
  year={2017}
}

@article{yao2024trajclip,
  title={TrajCLIP: Pedestrian trajectory prediction method using contrastive learning and idempotent networks},
  author={Yao, Pengfei and Zhu, Yinglong and Bi, Huikun and Mao, Tianlu and Wang, Zhaoqi},
  journal={Advances in Neural Information Processing Systems},
  volume={37},
  pages={77023--77037},
  year={2024}
}

@inproceedings{chen2025socialmoif,
  title={SocialMOIF: Multi-Order Intention Fusion for Pedestrian Trajectory Prediction},
  author={Chen, Kai and Zhao, Xiaodong and Huang, Yujie and Fang, Guoyu and Song, Xiao and Wang, Ruiping and Wang, Ziyuan},
  booktitle={Proceedings of the Computer Vision and Pattern Recognition Conference},
  pages={22465--22475},
  year={2025}
}

@inproceedings{wang2024driving,
  title={Driving into the future: Multiview visual forecasting and planning with world model for autonomous driving},
  author={Wang, Yuqi and He, Jiawei and Fan, Lue and Li, Hongxin and Chen, Yuntao and Zhang, Zhaoxiang},
  booktitle={Proceedings of the IEEE/CVF Conference on Computer Vision and Pattern Recognition},
  pages={14749--14759},
  year={2024}
}

@InProceedings{Ham_2023_CVPR,
    author    = {Ham, Je-Seok and Kim, Dae Hoe and Jung, NamKyo and Moon, Jinyoung},
    title     = {CIPF: Crossing Intention Prediction Network Based on Feature Fusion Modules for Improving Pedestrian Safety},
    booktitle = {Proceedings of the IEEE/CVF Conference on Computer Vision and Pattern Recognition (CVPR) Workshops},
    month     = {June},
    year      = {2023},
    pages     = {3666-3675}
}

@inproceedings{malla2020titan,
  title={TITAN: Future Forecast using Action Priors},
  author={Malla, Srikanth and Dariush, Behzad and Choi, Chiho},
  booktitle={Proceedings of the IEEE/CVF Conference on Computer Vision and Pattern Recognition},
  pages={11186--11196},
  year={2020}
}

@inproceedings{rasouli2017ICCVW,
title={Are they going to cross? A benchmark dataset and baseline for pedestrian crosswalk behavior},
author={Rasouli, Amir and Kotseruba, Iuliia and Tsotsos, John K},
booktitle={Proceedings of the IEEE International Conference on Computer Vision Workshops},
pages={206--213},
year={2017}
}

@inproceedings{Rasouli2017IV,
title={Agreeing to cross: How drivers and pedestrians communicate},
author={Rasouli, Amir and Kotseruba, Iuliia and Tsotsos, John K},
booktitle={IEEE Intelligent Vehicles Symposium (IV)},
pages={264--269},
year={2017}
}

@inproceedings{Rasouli2019PIE,
author = {Amir Rasouli and Iuliia Kotseruba and Toni Kunic and John K. Tsotsos},
title = {PIE: A Large-Scale Dataset and Models for Pedestrian Intention Estimation and Trajectory Prediction},
booktitle = {International Conference on Computer Vision (ICCV)},
year = {2019} }

@ARTICLE{Munir:VLM:2025,
  author={Munir, Farzeen and Azam, Shoaib and Mihaylova, Tsvetomila and Kyrki, Ville and Kucner, Tomasz Piotr},
  journal={IEEE Open Journal of Intelligent Transportation Systems}, 
  title={Pedestrian Vision Language Model for Intentions Prediction}, 
  year={2025},
  volume={6},
  number={},
  pages={393-406},
  doi={10.1109/OJITS.2025.3554387}}

@INPROCEEDINGS{Crosato:ICHMS2021,
  author={Crosato, Luca and Wei, Chongfeng and Ho, Edmond S. L. and Shum, Hubert P. H.},
  booktitle={2021 IEEE 2nd International Conference on Human-Machine Systems (ICHMS)}, 
  title={Human-centric Autonomous Driving in an AV-Pedestrian Interactive Environment Using SVO}, 
  year={2021},
  volume={},
  number={},
  pages={1-6},
  doi={10.1109/ICHMS53169.2021.9582640}}

@article{Crosato:review,
author = {Crosato, Luca and Tian, Kai and Shum, Hubert P. H. and Ho, Edmond S. L. and Wang, Yafei and Wei, Chongfeng},
title = {Social Interaction-Aware Dynamical Models and Decision-Making for Autonomous Vehicles},
journal = {Advanced Intelligent Systems},
volume = {6},
number = {3},
pages = {2300575},
doi = {https://doi.org/10.1002/aisy.202300575},
year = {2024}
}

@inproceedings{Crosato:VR,
author = {Crosato, Luca and Wei, Chongfeng and Ho, Edmond S. L. and Shum, Hubert P. H. and Sun, Yuzhu},
title = {A Virtual Reality Framework for Human-Driver Interaction Research: Safe and Cost-Effective Data Collection},
year = {2024},
isbn = {9798400703225},
publisher = {Association for Computing Machinery},
address = {New York, NY, USA},
url = {https://doi.org/10.1145/3610977.3634923},
doi = {10.1145/3610977.3634923},
booktitle = {Proceedings of the 2024 ACM/IEEE International Conference on Human-Robot Interaction},
pages = {167–174},
numpages = {8},
location = {Boulder, CO, USA},
series = {HRI '24}
}

\end{document}